\documentclass[a4paper,fleqn]{cas-dc}

\usepackage[numbers]{natbib}
\usepackage{amsmath,amsfonts}
\usepackage{array}
\usepackage[caption=false,font=normalsize,labelfont=sf,textfont=sf]{subfig}
\usepackage{bm}
\usepackage{stfloats}
\usepackage{amsthm}
\usepackage{algorithm}
\usepackage{algorithmic}
\usepackage{bbding}
\usepackage{pifont}

\usepackage{amssymb}
\usepackage{makecell}
\usepackage{color}
\usepackage{caption}
\usepackage{xcolor}
\usepackage{multicol}
\usepackage{hyperref}
\usepackage{multirow}
\usepackage{booktabs}
\usepackage{textcomp}
\usepackage{stfloats}
\usepackage{url}
\usepackage{verbatim}
\usepackage{graphicx}
\usepackage{subcaption}
\usepackage{hyperref}

\begin{document}
\let\WriteBookmarks\relax
\def\floatpagepagefraction{1}
\def\textpagefraction{.001}

\shortauthors{Liao et~al.}

\title[mode = title]{SafeCast: Risk-Responsive Motion Forecasting for Autonomous Vehicles}



\author[1,2]{Haicheng Liao}
\fnmark[1]
\credit{Conceptualization, Methodology, Experiment, Writing}

\author[1]{Hanlin Kong}
\credit{Experiment}
\fnmark[1]

\author[1]{Bin Rao}
\credit{Methodology, Writing}

\author[1,2]{Bonan Wang}
\credit{Methodology}

\author[1,3]{Chengyue Wang}
\credit{Experiment}

\author[1]{Guyang Yu}
\credit{Experiment}

\author[1,2]{Yuming Huang}
\credit{Methodology}

\author[1,3]{Ruru Tang}
\credit{Methodology}

\author[1,2]{Chengzhong Xu}
\credit{Methodology}

\author[1,2,3]{Zhenning Li}
\cormark[1]
\ead{zhenningli@um.edu.mo}
\credit{Conceptualization, Review}

\affiliation[1]{organization={State Key Laboratory of Internet of Things for Smart City, University of Macau}, city={Macau SAR}, country={China}}

\affiliation[2]{organization={Department of Computer and Information Science, University of Macau}, city={Macau SAR}, country={China}}

\affiliation[3]{organization={Department of Civil and Environmental Engineering, University of Macau}, city={Macau SAR}, country={China}}

\cortext[cor1]{Corresponding author; $^{1}$Equally Contributed}
\fntext[fn1]{https://orcid.org/0000-0002-0877-6829 (Z. Li)}

\begin{abstract}
Accurate motion forecasting is essential for the safety and reliability of autonomous driving (AD) systems. While existing methods have made significant progress, they often overlook explicit safety constraints and struggle to capture the complex interactions among traffic agents, environmental factors, and motion dynamics. To address these challenges, we present SafeCast, a risk-responsive motion forecasting model that integrates safety-aware decision-making with uncertainty-aware adaptability. SafeCast is the first to incorporate the Responsibility-Sensitive Safety (RSS) framework into motion forecasting, encoding interpretable safety rules—such as safe distances and collision avoidance—based on traffic norms and physical principles. To further enhance robustness, we introduce the Graph Uncertainty Feature (GUF), a graph-based module that injects learnable noise into Graph Attention Networks, capturing real-world uncertainties and enhancing generalization across diverse scenarios. We evaluate SafeCast on four real-world benchmark datasets— Next Generation Simulation (NGSIM), Highway Drone (HighD), ApolloScape, and the Macao Connected Autonomous Driving (MoCAD)—covering highway, urban, and mixed-autonomy traffic environments. Our model achieves state-of-the-art (SOTA) accuracy while maintaining a lightweight architecture and low inference latency, underscoring its potential for real-time deployment in safety-critical AD systems.
\end{abstract}

\maketitle

\section{Introduction}\label{Indro}
The field of autonomous driving (AD) is undergoing rapid advancements, confronting the complex challenge of accurately forecasting vehicle motion. This task, extending beyond mere path plotting, involves intricate decision-making influenced by environmental conditions and a steadfast commitment to safety \citep{geng2023physics,li2024steering}. Understanding these variables is essential for developing models that can navigate the dynamic nature of real-world driving.

Existing approaches\cite{chen2023stochastic,liao2024cdstraj,liao2024less,wang2025nest} to vehicle motion forecasting have predominantly hinged on the analysis of historical states, leveraging observed patterns to extrapolate future positions. However, these methodologies \cite{chen2022intention,wang2025wake,liao2024gpt} often prove inadequate for incorporating safety considerations, which are of paramount importance in real-world traffic environments. A common limitation is their inability to explicitly model safety metrics, such as maintaining appropriate longitudinal and lateral distances, which are essential for collision avoidance. For instance, many models lack mechanisms to enforce collision-avoidance principles like safe longitudinal and lateral distances \cite{liao2024Trajectory,wang2025dynamics} or fail to reconcile safety constraints with the complex interplay of agent intentions, kinematics, and environmental variables \cite{liao2024gpt}. Consequently, these methods risk generating predictions that are statistically plausible yet physically unsafe, particularly in edge cases such as abrupt lane changes, intersections with limited visibility, or high-density traffic scenarios. Moreover, these models frequently fail to account for the intricate interrelationships between safety constraints, motion dynamics, and environmental variables, thereby compromising their resilience and applicability in uncertain or high-risk scenarios \cite{liao2024crash}.

\begin{figure}
\centering
\includegraphics[width=0.48\textwidth]{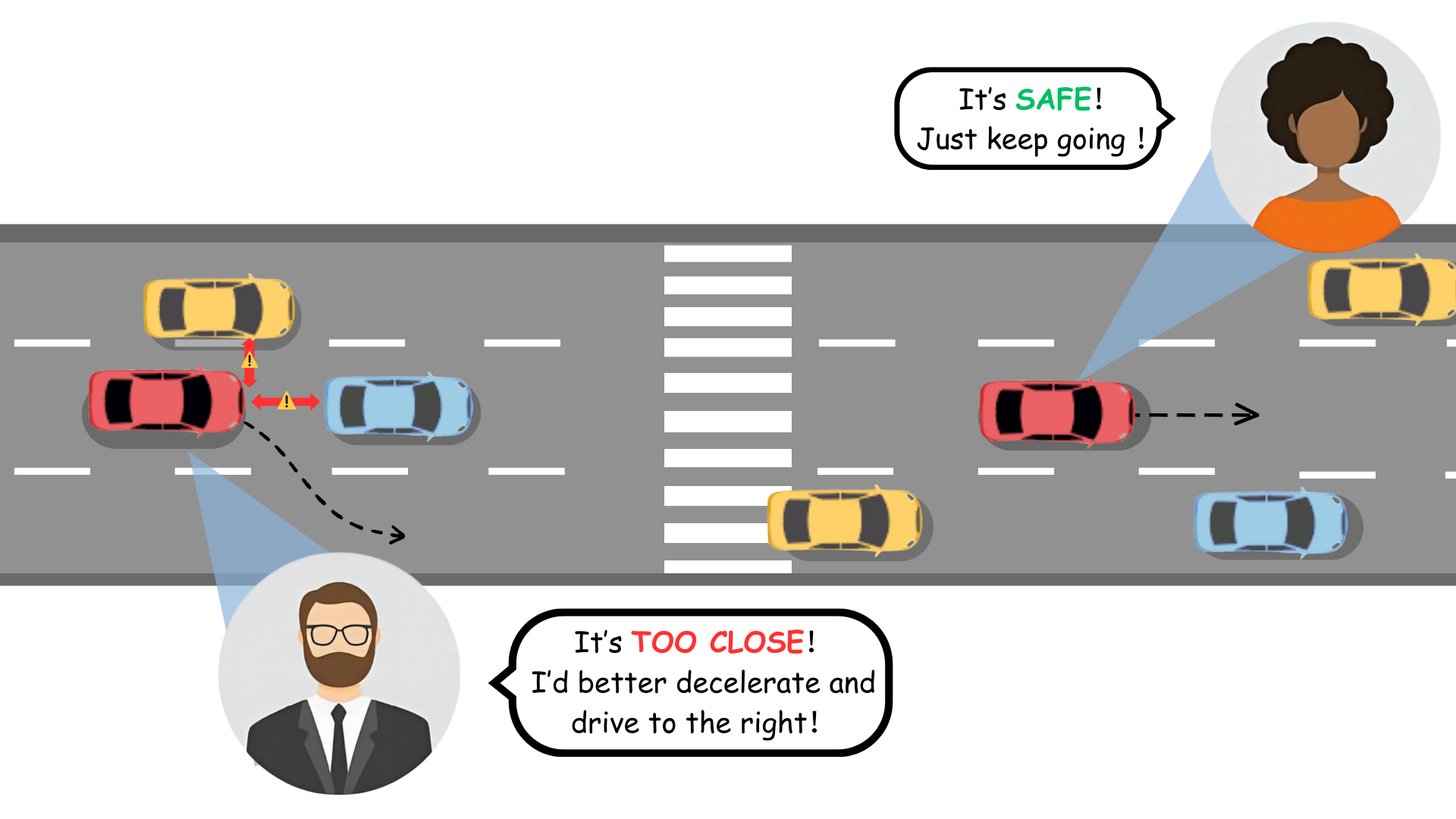}
\caption{Illustrative representation of safety considerations impacting the decision-making process and subsequent motion of traffic agents.}
\label{fig:head}
\end{figure}

{
Recognizing the critical need for enhanced safety and adaptability in forecasting, we introduce \textbf{SafeCast}, a risk-responsive motion forecasting framework that jointly considers safety-aware decision-making and environment-driven uncertainty. As shown in Fig.~\ref{fig:head}, our key insight is to model vehicle motion not only as a physical prediction problem but also as a structured decision-making process, where each motion trajectory reflects rational and lawful driving behavior influenced by traffic rules and situational risks.

Central to the SafeCast model is the integration of the Responsibility-Sensitive Safety (RSS) framework \citep{shalev2017formal}, an innovative approach to autonomous driving safety proposed by Mobileye. The RSS framework offers a structured methodology to ensure safety through a series of formalized rules and logical constructs. These rules are derived from established driving principles and physics, covering crucial elements like maintaining safe distances, adhering to the right of way, managing limited visibility conditions, and responding effectively in potential collision scenarios. Integrating RSS enables our model to interpret vehicle motion as a sequence of safe, deliberate choices made by the driver, deeply embedding a safety-first philosophy into the core of motion forecasting. By embedding RSS as prior domain knowledge, SafeCast significantly enhances the safety robustness of motion forecasting.

To further enhance adaptability and robustness, SafeCast incorporates the Graph Uncertainty Feature (GUF), inspired by the diffusion model \citep{ho2020denoising} and Graph Prompt Feature \citep{fang2023universal}. The GUF infuses variable-learnable noise into the Graph Attention Network, markedly boosting the model's ability to adjust to the diverse and unpredictable nature of real-world traffic. This feature is vital in addressing common challenges in motion forecasting, such as overfitting and representing complex driving environments accurately. Overall, the primary contributions of this study are summarized as follows:

\begin{itemize}
\item We propose SafeCast, the first motion forecasting framework to integrate the RSS safety model, ensuring that predictions are in line with realistic driving behaviors and stringent safety standards.

\item We introduce GUF, a graph-based uncertainty modeling technique that enhances the model’s robustness by simulating agent-level variability through learnable noise injection, leading to better generalization and more precise predictions in intricate driving scenes.

\item We evaluate SafeCast on four datasets—NGSIM, HighD, ApolloScape, and MoCAD—covering a wide range of highway and urban driving scenarios. SafeCast achieves state-of-the-art performance in accuracy and robustness while maintaining low computational overhead, demonstrating its robustness, efficiency, and practical applicability in AD systems.

\end{itemize}
}
This paper is organized as follows: Section~\ref{related_work} reviews related work. Section~\ref{method} presents the problem formulation and SafeCast architecture. Section~\ref{sec:5} reports experimental results and analysis. Section~\ref{sec:6} concludes with a summary and implications for autonomous vehicle (AV) technology.

\section{Related Work} \label{related_work}

\subsection{Motion Forecasting for Autonomous Vehicles}
A notable benchmark in motion forecasting is the CS-LSTM model \citep{deo2018convolutional}, which introduces a social grid framework to capture spatiotemporal interactions among traffic agents. This approach exemplifies the effective use of CNNs \citep{Wang_Wang_Yan_Wang_2023,lu20213d} and RNNs \citep{lai2024drive} in handling multimodal trajectory prediction. Building upon this foundation, subsequent research has explored advanced architectures to improve prediction accuracy and contextual reasoning. These include Generative Adversarial Networks (GANs) \citep{zhao2019multi,zhou2023query}, attention-based models \citep{hybrid,SFM,li2023context,zhang2022explainable}, and transformer-based frameworks \citep{Zhou_2022_CVPR,li2023weakly,huang2022multi,mo2023map}. For instance, models such as STDAN \citep{chen2022intention} and GAVA \citep{liao2024human} leverage spatio-temporal attention to model interactions effectively. Similarly, FHIF \citep{zuo2023trajectory} and MHA-LSTM \citep{messaoud2020attention} employ multi-head attention to improve temporal reasoning. Recent innovations include DACR-AMTP \citep{cong2023dacr}, which utilizes a drivable area class graph to enhance spatial constraints, and iNATran \citep{cheng2022gatraj}, a transformer-based model tailored for interpretable forecasting. BAT \citep{liao2024bat} incorporates human psychological factors and decision-making uncertainty, while HLTP \citep{10468619-HLTP} and HLTP++ \citep{liao2024less} leverage human visual perception for dynamic scene modeling. Graph-based models such as MFTraj \citep{ijcai2024p0657} and AI-TP \citep{zhang2022ai} balance computational efficiency with accuracy by explicitly modelling agent interactions. Furthermore, recent works have begun to explore the integration of Large Language Models (LLMs) \citep{liao2025cot} to improve scene understanding and reasoning capabilities in trajectory prediction. Despite these advancements, many deep learning-based methods remain limited by their lack of interpretability and high computational cost. Incorporating domain knowledge and physics-informed constraints offers a promising direction for enhancing transparency, safety awareness, and computational efficiency in motion forecasting for autonomous driving systems.

\subsection{Safety Metrics in Motion Forecasting}
{
Safety is a foundational requirement in motion forecasting for autonomous driving, necessitating the integration of robust and interpretable safety metrics. Among the most widely recognized are the Responsibility-Sensitive Safety metric \citep{shalev2017formal} and the Safe Force Field (SFF) model \citep{nister2019safety}. RSS provides a mathematically grounded, rule-based framework rooted in kinematic principles, offering explicit guidelines for maintaining safe following distances, lateral spacing, and right-of-way. This level of interpretability ensures a transparent evaluation of safety in diverse traffic scenarios and supports formal compliance with safety standards. In contrast, the SFF model employs a dynamic force field to assess safety, focusing primarily on maintaining safe distances between traffic agents. While both metrics aim to enhance safety, the RSS metric's rigorously defined, physics-based principles make it particularly attractive for ensuring compliance with established safety standards. Its interpretability and clarity offer a distinct advantage, particularly when integrating safety into motion forecasting models for autonomous systems. Recent studies, such as DACR-AMTP \citep{cong2023dacr}, CDTP \cite{ijcai2024p656}, CITF \cite{liao2025minds}, and CF-LSTM \citep{xie2021congestion} also incorporate safety considerations by modeling drivable areas, congestion patterns, or interaction intensity \citep{huang2024dtpp}. However, these approaches often lack the explicitness and generalizability of RSS. Its ability to quantify safety through interpretable kinematic constraints continues to make it a compelling choice for designing safe, collision-free behavior for AVs.}

\begin{figure*}
\centering
\includegraphics[width=0.9\textwidth]{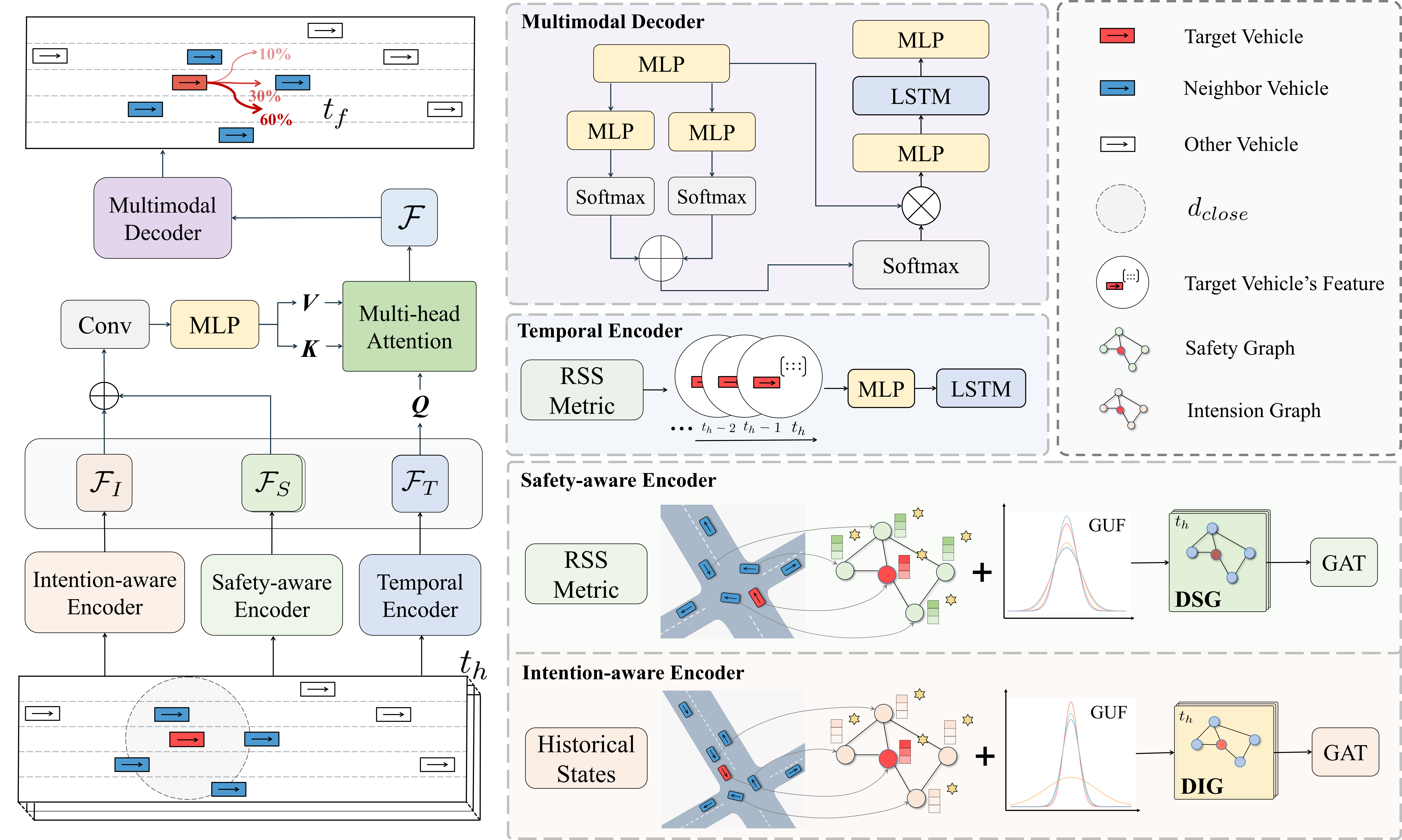}
\caption{An overview of the architecture of our proposed model, which includes the intention-aware module, the safety-aware module, the temporal encoder, and the multimodal decoder. The overall process of SafeCast is depicted from bottom to top.}
\label{fig:structure}
\end{figure*}

\section{Methodology}\label{method}
\subsection{Problem Formulation}
The primary goal of our research is to forecast the motion of the \textit{target vehicle}, encompassing all traffic agents within the sensing range of the AV in a mixed autonomy environment. Given the observed historical state $\bm{X}_{0:n}^{t-t_{h}:t}$ of the target vehicle (denoted as superscript 0) and its surrounding traffic agents (superscripts from 1 to $n$) over the time interval from $t-t_{h}$ to $t$, the task is to predict the future motion of the target vehicle $\bm{Y}$ over a specified prediction horizon $t_{h}$.
Moreover, at time step $t$, the observed historical state $\bm{X}_{0:n}^{t-t_{h}:t}$ comprises a series of vectors spanning the past $t_h$ time steps. These vectors encompass position coordinates $\{\bm{p}_{0: n}^{t-t_{h}:t}\}$, instant velocity $\{\bm{v}_{0: n}^{t-t_{h}:t}\}$, acceleration $\{\bm{a}_{0: n}^{t-t_{h}:t}\}$ and vehicle type $\{\bm{u}_{0: n}^{t-t_{h}:t}\}$. All these vectors form the input for our proposed model, providing a 
comprehensive data foundation for motion forecasting.

\subsection{Overview}
The overall architecture of SafeCast is depicted in Fig.~\ref{fig:structure}. The model adopts an encoder-decoder framework comprising four main components: the \textbf{intention-aware module}, \textbf{safety-aware module}, \textbf{feature fusion module}, and \textbf{multimodal decoder}. Each component is designed to capture distinct yet complementary aspects of traffic dynamics, contributing to accurate and efficient motion forecasting for autonomous vehicles. Unlike deterministic forecasting approaches, SafeCast formulates motion prediction as a probabilistic estimation task. Given the historical trajectories of the ego vehicle and surrounding agents, denoted as $\bm{X}_{0:n}^{t-t_{h}:t}$, the model estimates a conditional probability distribution over future trajectories, expressed as $P(\bm{Y}|\bm{X}_{0:n}^{t-t_{h}:t})$. This formulation allows SafeCast to account for inherent uncertainty and behavioral variability, producing a multimodal output that captures diverse future maneuvers. A detailed overview of each module is presented in the following section.

\subsection{Intention-aware Module}
This module is designed to tackle the challenges of motion forecasting in highly interactive and dynamic traffic environments. Its core objective is to effectively capture spatiotemporal dependencies and model the uncertainty inherent in agent interactions while maintaining robustness and generalizability to previously unseen scenarios. To this end, we propose two key components: the {Dynamic Intention Graph (DIG)} and the {Uncertainty-aware Graph Attention Network (Uncertainty-aware GAT)}. Together, these modules empower the SafeCast model to reason over complex agent behaviors and adapt to diverse traffic conditions, significantly improving its performance in real-world applications.

\subsubsection{Dynamic Intention Graph}
The DIG is constructed to represent a traffic scene at a specific time step, encompassing the real-time states of the target vehicle and its surrounding agents to represent a traffic scene at a specific time step. At a given time $t_k$, the DIG, denoted as $\mathcal{G}_{\textit{DIG}}^{t_k}$ can be formulated:
\begin{equation}\label{eq.7-1}
\mathcal{G}_{\textit{DIG}}^{t_k}= \{\mathcal{V}^{t_k},\mathcal{E}^{t_k}\}
\end{equation}
{where $ \mathcal{V}^{t_{k}}=\{ v_{i}^{t_k}, \,i \in [0, n]\} $ is the node set. Each node \( {v}_{i}^{t_k} \) corresponds to a traffic agent and is characterized by its historical states $\bm{X}_{i}^{t_{k}}=\{\bm{p}_{i}^{t_{k}}, \bm{v}_{i}^{t_{k}}, \bm{a}_{i}^{t_{k}}, \bm{u}_{i}^{t_{k}}\}$.} Formally, the edge set ${\mathcal{E}^{t_{k}}} = \{{e_{0}^{t_{k}}},{e_{1}^{t_{k}}}\ldots,{e_{n}^{t_{k}}}\}$ represents the potential interactions between agents, where each edge ${e_{i}^{t_{k}}}=\{v_{i}^{t_k} v_{j}^{t_k}\mid (i, j) \in \mathcal{D}^{t_k}, i \neq j\}$ connects the node $v_{i}^{t_k}$ with other agents that have potential influences with it. Moreover, $\mathcal{D}^{t_k}$ is the set of agents in proximity at time $t_k$, defined by a shortest distance $d\left(v_{i}^{t}, v_{j}^{t}\right)$ less than or equal to a pre-established threshold $d_{close}$.  Additionally, the adjacency matrix $\mathcal{A}^{t_k} \in \mathbb{R}^{n \times n}$ for the DIG $\mathcal{G}_{\textit{DIG}}^{t_k}$ at time $t_k$ can be defined as follows:
\begin{equation}
\mathcal{A}^{t_k}(i, j)=
\begin{cases}1 & \text { if } d\left(v_{i}^{t_k}, v_{j}^{t_k}\right)\leq{d_{close}}, \\ 0 & \text {otherwise}
\end{cases}
i \neq j, \, i, j \in [0,n]
\end{equation}
Next, the DIG $\mathcal{G}_{\textit{DIG}}^{t-t_{h}:t}$ is fed into our proposed uncertainty-aware GAT  to capture the dynamic interactions and influences among different agents in a traffic scene, resulting in the intention-aware features $\mathcal{F}_{I}$.

\subsubsection{Uncertainty-aware Graph Attention Network}
\begin{figure*}
\centering
\includegraphics[width=0.91\textwidth]{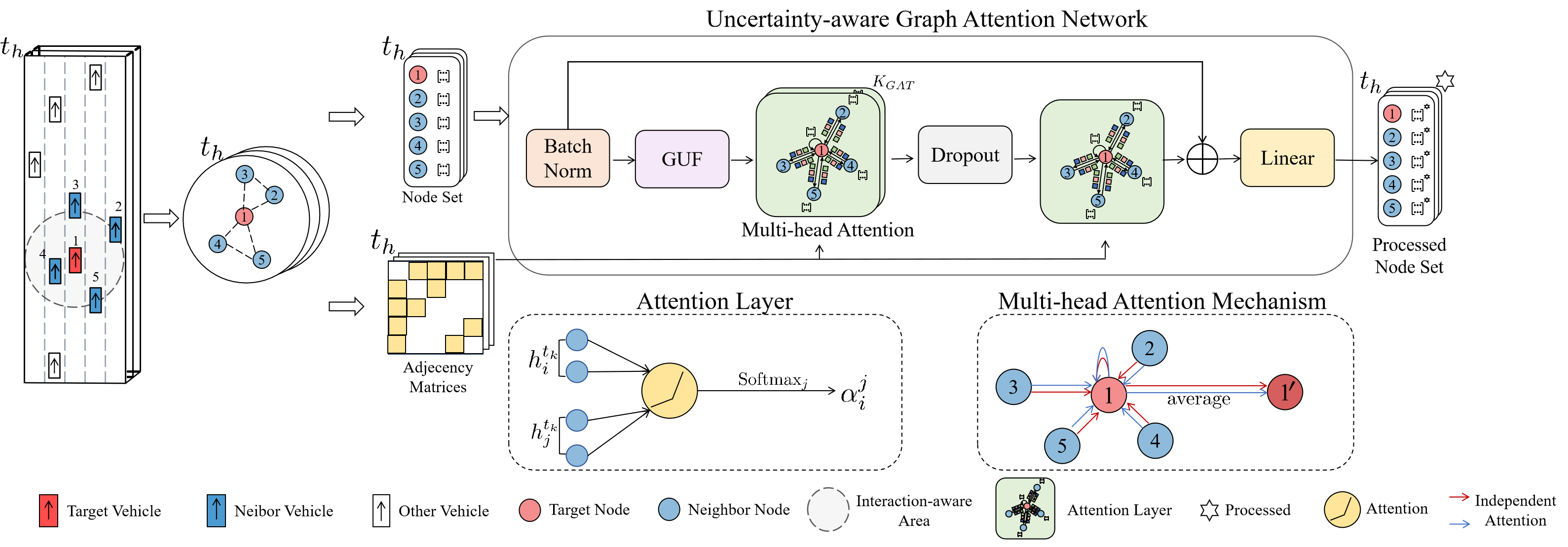}
\caption{{Illustration of the architecture of the uncertainty-aware GAT. It consists of a batch-normalization layer, a GUF layer, $K_{GAT}$ attention layers for multi-head attention, a dropout layer, another attention layer, and a linear layer. The GUF layer enables the network to estimate and consider the uncertainty of the node features, enhancing its robustness to noisy or incomplete data.}}
\label{fig:gat}
\end{figure*}

We propose a novel framework, the Uncertainty-aware GAT, as shown in Fig. \ref{fig:gat}, to account for the variables and uncertainties in real-world driving scenes. {Traditional graph attention mechanisms assume deterministic relationships, which limits their ability to generalize in dynamic and unpredictable environments. To overcome this, we integrate a GUF mechanism into the GAT framework.} GUF introduces variance-learnable Gaussian noise and a shared linear transformation to realistically simulate unpredictable elements in driving, such as human behavior and potential vehicle maneuvers. This can be represented as follows:
\begin{equation}
    \begin{cases}
         \mathcal{P} = \{p_1, p_2, \cdots, p_F\}, \quad
p_i \sim \mathcal{N}(\mu, \sigma_i^2) \\
    h^{t_k}_i = \mathbf{W} (v_i^{t_k} + \mathcal{P})
    \end{cases}
\end{equation}
Here, $\mathcal{P}$ represents a set of Gaussian noise components, each following a distribution $\mathcal{N}(\mu, \sigma_i^2)$ with mean $\mu$ and learnable standard deviation $\sigma_i$, where the weight is shared across all nodes at all time steps, and $F$ is the feature dimension of $v_i^{t_k}$. Furthermore, the transformed node features $h^{t_k}_i$ is obtained by applying the shared weight matrix $\mathbf{W}$ to the sum of the node features $v_i^{t_k}$ and the noise components.  
{The attention mechanism quantifies the influence between agents while integrating the uncertainty-enhanced features. The attention coefficient $\gamma_i^j$ is computed as follows:}
\begin{equation}
    \gamma_i^j = \phi_{ \textit{LR}}\left(\mathbf{a}^{\top}(h^{t_k}_i\|h^{t_k}_j)\right)
\end{equation}
where $\phi_{ \textit{LR}}$ refers to the Leaky Rectified Linear Unit activation feature, and $\|$ represents the concatenation operation. We implement a masked attention mechanism for preserving vital structural information and concentrate solely on relevant neighboring nodes.  The masked attention coefficients $\alpha_i^j$ are computed through the softmax activation feature:
\begin{equation}
    \alpha_i^j =  \phi_{ \textit{Softmax}}(\gamma_i^j) = \frac{e^{\gamma_i^j}} {\sum_{n\in \mathcal{N}_i}{e^{\gamma_i^n}}}
\end{equation}
{where $\phi_{ \textit{Softmax}}$ denotes the softmax activation feature, while $\mathcal{N}_i$ represents the neighbors of traffic agent $v_i$. To improve learning stability and model capacity, multi-head attention aggregates information from multiple perspectives, and residual connections facilitate gradient flow. Specifically, $K_{GAT}$ GAT layers perform attention mechanisms, and their outputs are averaged to obtain the output features:}
\begin{equation}
    \hat{v}^{t_k}_i = \phi_{ \textit{ELU}}(\sum_{head=1}^{K_{GAT}}\frac{\sum_{j\in \mathcal{N}_i}\alpha_i^j \cdot h^{t_k}_j}{K_{GAT}}) 
    \label{eq:gat}
\end{equation}
where $\phi_{ \textit{ELU}}$ represents the Exponential Linear Unit (ELU) activation function. {The output features are processed through a Dropout layer to mitigate overfitting. Next, an additional GAT layer captures high-level interactions among traffic agents. Residual connections are integrated into the architecture to enhance gradient flow and improve training stability. Finally, the refined features are passed through a linear layer for reshaping into the desired format.}

\subsection{Safety-Aware Module}
{Safety is a fundamental aspect of AV motion prediction. Maintaining safe distances between traffic agents ensures reliable and collision-free motion planning, even in complex or dynamic scenarios. To achieve this, the safety-aware module integrates safety distance metrics derived from the RSS framework into the prediction task. These metrics serve as input features, providing the model with prior knowledge of safe distances that guide the prediction process. The module enhances the model’s ability to generalize to diverse traffic conditions, including extreme or unseen scenarios, by embedding these physics-based principles.
\subsubsection{Safety Distance Metric}
\begin{figure}
\centering
\includegraphics[width=0.49\textwidth]{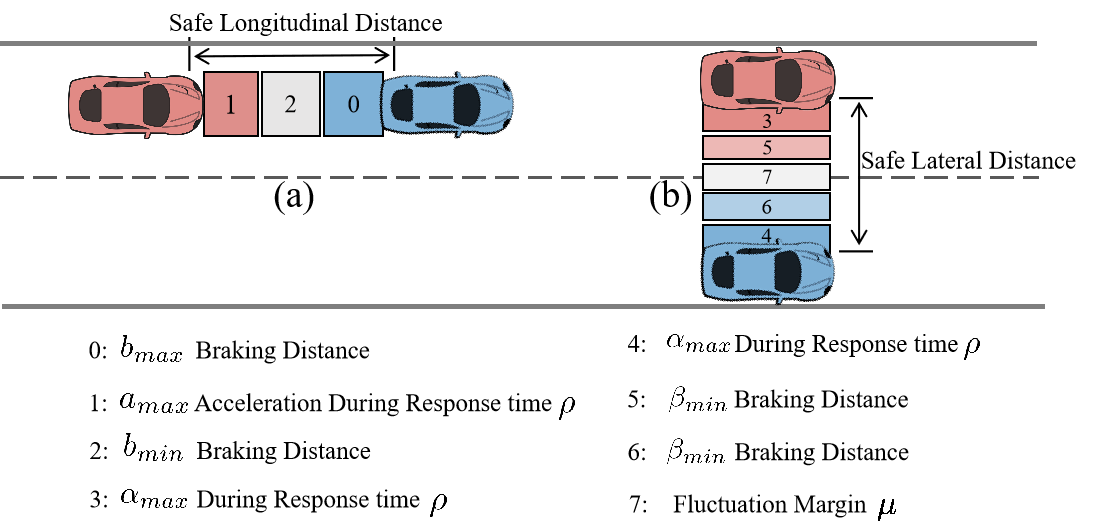}
\caption{Illustration of safe longitudinal and lateral distance metrics in the RSS framework: (a) Safe longitudinal distance, accounting for interactions within the same lane. (b) Safe lateral distance, addressing interactions in adjacent lanes.}
\label{fig:rss}
\end{figure}

Drawing from foundational physics and practical driving experience, novice drivers are typically advised to maintain a safe following distance to allow sufficient reaction time in case the lead vehicle decelerates unexpectedly. In autonomous driving systems, this principle is operationalized through the definition of safety-aware longitudinal and lateral distances. These metrics formalize the minimum safe spacing required to prevent collisions, enabling the AV to react appropriately to dynamic changes in the environment. This subsection defines these distances precisely.

\textbf{Safe Longitudinal Distance.}
To ensure safety between two vehicles traveling in the same direction, the minimum safe longitudinal distance is defined as follows:
\begin{equation}
\label{eq:rss1}
    d^{\text{lon}}_{\min} = \left[v_r \rho + \frac{1}{2} a_{\max} \rho^2 + \frac{(v_r + \rho a_{\max})^2}{2 b_{\min}} - \frac{v_f^2}{2 b_{\max}}\right]_+
\end{equation}
where $d^{\text{lon}}_{\min}$ denotes the minimum required longitudinal distance to avoid a rear-end collision. The operator $[\cdot]_+$ ensures non-negativity by returning the value itself if positive or zero otherwise. $v_f$ and $v_r$ are the longitudinal velocities of the front and rear vehicles, respectively. $\rho$ is the driver (or system) reaction time, $a_{\max}$ is the maximum acceleration, while $b_{\max}$ and $b_{\min}$ represent the maximum and minimum braking decelerations of the front and rear vehicles, respectively.

\textbf{Safe Lateral Distance.}
To maintain a safe buffer between adjacent vehicles, the minimum lateral distance can be formally defined as follows:
\begin{equation}
\label{eq:rss2}
\begin{cases}
    v_{1,\rho} = v_1 + \alpha_{\max} \rho \\
    v_{2,\rho} = v_2 + \alpha_{\max} \rho \\
    d^{\text{lat}}_{\min} = \mu + \left[ \frac{(v_1 + v_{1,\rho})}{2} \rho + \frac{v_{1,\rho}^2}{2 \beta_{\min}} - \left( \frac{v_2 + v_{2,\rho}}{2} \rho - \frac{v_{2,\rho}^2}{2 \beta_{\min}} \right) \right]_+
\end{cases}
\end{equation}

Here, $v_1$ and $v_2$ are the lateral velocities of the right and left vehicles, respectively, and $\rho$ is the reaction time. $\alpha_{\max}$ denotes the maximum lateral acceleration, while $\beta_{\min}$ is the minimum lateral deceleration. $\mu$ accounts for the necessary lateral safety margin due to natural fluctuations. The operator $[\cdot]_+$ ensures non-negativity. By incorporating these safety-aware metrics directly into the model architecture, SafeCast avoids reliance on post hoc rule-based constraints. Unlike traditional motion planners that apply safety checks after prediction, SafeCast embeds interpretable physical and legal constraints into its learning process. This allows the model to make predictions that are inherently safety-compliant, improving generalization across dense traffic, unstructured road layouts, and edge-case scenarios. 

The RSS framework provides a structured, interpretable foundation for motion safety. Its parameters—based on vehicular dynamics and empirical traffic standards—offer consistent safety margins across diverse driving conditions. These fixed settings reduce overfitting, simplify the learning process, and enhance transferability. Furthermore, the mathematical flexibility of RSS allows for future integration with real-time perception and adaptive control modules, enabling dynamic tuning of parameters. This positions RSS not only as a conservative safety baseline but also as a scalable safety prior for real-world autonomous driving systems.}

\subsubsection{Spatial Encoder}
To capture safety-critical spatial interactions between the target vehicle and its surrounding agents, we propose the {Dynamic Safety Graph (DSG)}—a novel graph structure that integrates RSS distance metrics into its representation. At each time step \( t_k \), the DSG is defined as an undirected graph \( \tilde{G}^{t_k} = \{\tilde{V}^{t_k}, \tilde{E}^{t_k}\} \), where the node set \( \tilde{V}^{t_k} = \{ \tilde{v}_{i}^{t_k} \,|\, i \in [0, n] \} \) represents all traffic agents.
Each node \( \tilde{v}_{i}^{t_k} \) encodes the 2D position \( \bm{p}_{i}^{t_k} \), along with the agent’s dynamically computed maximum safe longitudinal and lateral distances \( d^{\text{lon}, t_k}_{\text{max}, i} \) and \( d^{\text{lat}, t_k}_{\text{max}, i} \), respectively:
\begin{equation}
\tilde{v}^{t_k}_{i} = \left[\bm{p}_{i}^{t_k}, d^{\text{lon}, t_k}_{\text{max}, i}, d^{\text{lat}, t_k}_{\text{max}, i} \right]^\top
\end{equation}

Correspondingly, edges \( \tilde{e}^{t_k}_{ij} \in \tilde{E}^{t_k} \) are created based on spatial proximity: an edge is established between nodes \( i \) and \( j \) if the Euclidean distance between them, \( d(\tilde{v}^{t_k}_{i}, \tilde{v}^{t_k}_{j}) \), is below a predefined threshold \( d^{\text{lon}}_{\text{close}} \). Safety distances are computed dynamically based on lane context—longitudinal safety for agents in the same lane and lateral safety for adjacent-lane interactions.
To effectively capture spatial dependencies under uncertainty, we apply the Uncertainty-aware GAT, as introduced in the intention-aware module. This enables the DSG to model both the physical configuration and the safety margin around each agent,

\subsubsection{Temporal Encoder}
To capture the temporal dependencies of the target vehicle, this encoder individually encodes pertinent features of the target vehicle, encompassing both the historical state and the safety distance metric. It applies a Multi-Layer Perceptron-ELU-LSTM structure.
Specifically,  the historical state of the target vehicle $\mathcal{T}_{0} =\{d^{lon,t_k}_{max, i}, d^{lat,t_k}_{max, i}, \bm{X}_{0}^{t-t_{h}:t} \} \in \mathbb{R}^{t_h \times {m}}$, where ${m}$ is the feature dimension. 
We utilize the MLP framework to extract and amplify features with an ELU activation function, followed by the LSTM network to capture feature variations over the time horizon $t_h$.
\begin{equation}
    \mathcal{F}_{T} = \phi_{\textit{LSTM}}\left(\phi_{\textit{ELU}}\left(\phi_{\textit{MLP}}(\mathcal{T}_{0}^{*}, W_{m}),W_{l}\right)\right)
\end{equation}
Here, $W_{m}$ and $W_{l}$ represent the learnable weight of the MLP and  LSTM, respectively. Moreover, the temporal encoder outputs the temporal feature of the target vehicle, denoted as the $\mathcal{F}_{T} \in \mathbb{R}^{\mathcal{B} \times t_h \times \mathcal{C}}$, where $\mathcal{B}$ is the batch size and $\mathcal{C}$ is the hyperparametric dimension for the LSTM network. 

\subsection{Feature Fusion Module}
\begin{figure}
\centering
\includegraphics[width=0.48\textwidth]{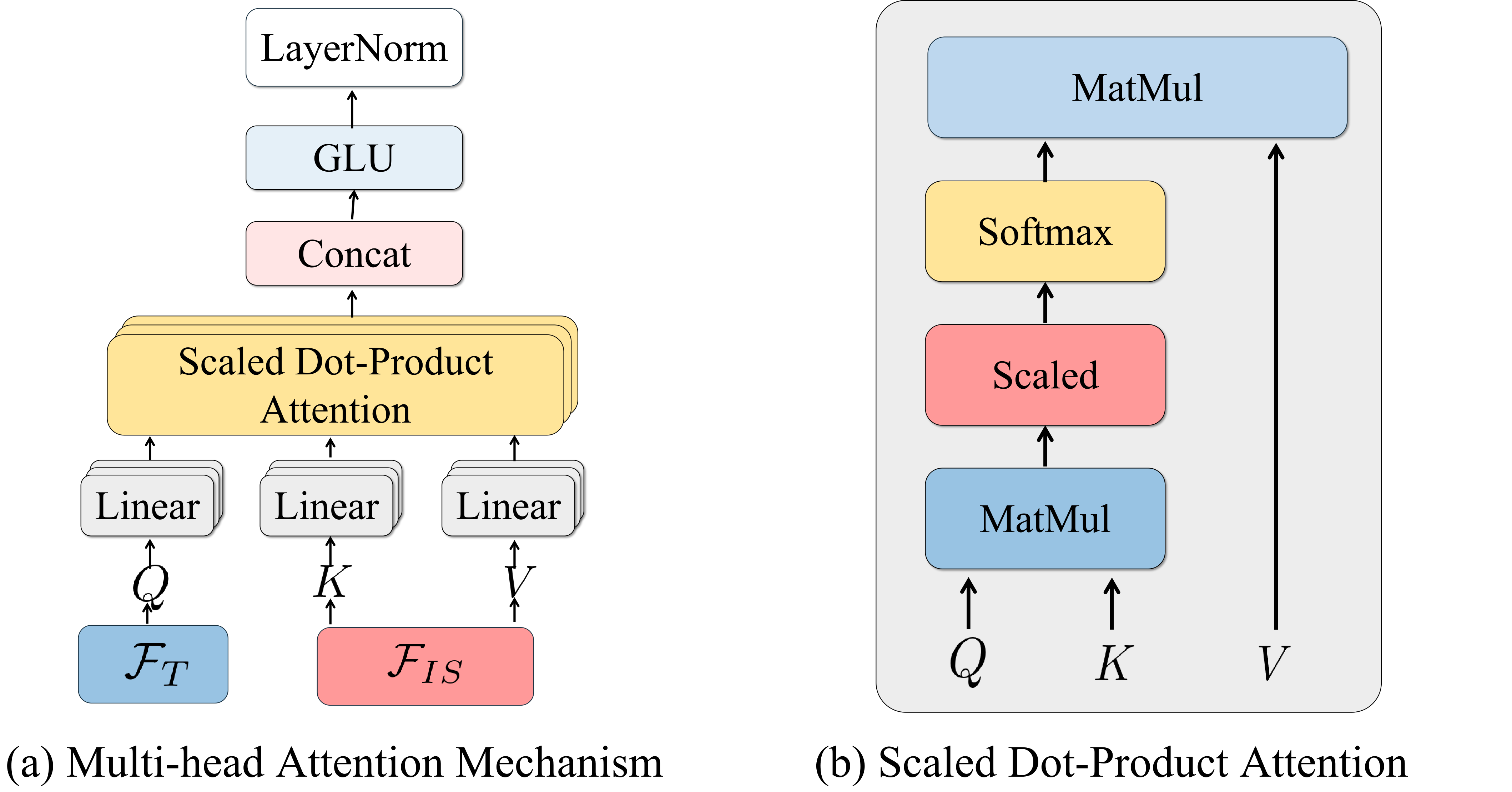}
\caption{Structure of the multi-head attention mechanism: (a) Integration of features $\mathcal{F}_{T}$ and $\mathcal{F}_{IS}$ through scaled dot-product attention. (b) Illustrates of weight computation.}
\label{fig:3}
\end{figure}

{The feature fusion module combines the outputs of the intention-aware $\mathcal{F}_{I}$ and the safety-aware $\mathcal{F}_{S}$ features using a multi-head scaled dot-product attention mechanism (Fig. \ref{fig:3}). The fused high-level feature $\mathcal{F}$ captures contextual and safety-critical information for downstream motion forecasting. Formally, the features $\mathcal{F}_{I}$ and $\mathcal{F}_{S}$ are concatenated and passed through a convolutional layer, ReLU activation, and an MLP for initial processing:}
\begin{equation}
        \mathcal{F}_{\textit{IS}} = \phi_{\textit{MLP}}\left(\phi_{\textit{ReLU}}(\phi_{\textit{Conv2D}}((\mathcal{F}_{I} \| \mathcal{F}_{S})))\right)
    \label{eq:conv}
\end{equation}
To enhance interaction modeling, the multi-head attention mechanism projects $\mathcal{F}_{T}$ and $\mathcal{F}_{\textit{IS}}$ into query $Q$, key $K$, and value $V$ vectors. Moreover, the scaled dot-product operation aggregates these projections across multiple heads to compute the attention output $H$:
\begin{equation}
\begin{cases}
H = \sum_{i=1}^{h} \alpha^{i} V^{i} \\
\alpha^{i} = \phi_{\textit{softmax}}\left(\frac{Q^{i}  K^{i}}{\sqrt{d_{K}}}\right)
\end{cases}
\label{eq:attention}
\end{equation}
Finally, the fused feature $\mathcal{F}$ undergoes further refinement via a Gated Linear Unit (GLU) and Layer Normalization (LN), ensuring efficient and stable feature extraction for multimodal decoding. Formally,
\begin{equation}
\mathcal{F}  = \phi_{\textit{LN}}\left(\phi_{\textit{GLU}}(H, W_{g})\right)
\end{equation}

\subsection{Multimodal Decoder}
{The multimodal decoder predicts the future motion of the target vehicle, accounting for uncertainty and multimodal behavior. The decoder employs a Gaussian Mixture Model (GMM) to generate lateral and longitudinal maneuver-based motion in a hierarchical framework:
\begin{equation}
\begin{cases}
M = \{M_{lat}, M_{lon}\} \\
M_{lat} = \{m_{S}, m_{R}, m_{L}\} \\
M_{lon} = \{m_{A}, m_{D}, m_{C}\}
\end{cases}
\label{eq:attention_1}
\end{equation}
where $M_{lat}$ includes lateral maneuvers (straight, right turn, left turn) and $M_{lon}$ includes longitudinal maneuvers (acceleration, deceleration, constant speed).} The probability distribution of the predicted motion $P(Y|{X}_{0:n}^{t-t_{h}:t})$ is computed:
\begin{equation}
    P(Y|{X}_{0:n}^{t-t_{h}:t}) = P_{\epsilon}(Y|{X}_{0:n}^{t-t_{h}:t}, M) P(M|{X}_{0:n}^{t-t_{h}:t})
\end{equation}

Here, $\epsilon$ denotes the parameters of the bivariate Gaussian distribution associated with the future motion of the target vehicle. It is obtained from the function $f_{dec}$ as follows:
\begin{equation}
    \epsilon = f_{dec}(\mathcal{F}\|P(M_{lat}|{X}_{0:n}^{t-t_{h}:t})\|P(M_{lon}|{X}_{0:n}^{t-t_{h}:t}))
\end{equation}
Notably, the function $f_{dec}$ is composed of two MLPs and one LSTM, which can be represented as follows:
\begin{equation}
    f_{dec}(\cdot)=\phi_{\textit{MLP}}(\phi_{\textit{LSTM}}[\phi_{\textit{MLP}}(\cdot)])
    \label{eq:decoder}
\end{equation}

\section{Experiments}\label{sec:5}
{
\subsection{Experiment Setup}
\noindent\textbf{Datasets.} We evaluate the effectiveness of our model on four real-world benchmarks that span diverse traffic complexities and environments, including highways, campuses, and urban streets: the Next Generation Simulation (NGSIM) dataset, the Highway Drone Dataset (HighD), ApolloScape, and the Macao Connected Autonomous Driving (MoCAD) dataset. These datasets provide time-stamped trajectories of traffic participants—including vehicles, pedestrians, and cyclists—represented in a local coordinate frame. Each dataset is partitioned into training, validation, and test sets to ensure experimental consistency and alignment with prior studies.

\noindent\textbf{Temporal Configuration.} For NGSIM, HighD, and MoCAD, trajectories are uniformly sampled over 8-second sequences. The initial 3 seconds (\(t_h = 3\)) serve as input history, while the following 5 seconds (\(t_f = 5\)) form the prediction horizon. This configuration enables robust evaluation in both highway and urban settings. Correspondingly, the ApolloScape benchmark follows the original challenge protocol: Predictions are made over a 3-second future window (\(t_f = 3\)) given a 3-second observation period (\(t_h = 3\)). This dataset introduces additional complexity with its inclusion of multiple agent types and diverse scene contexts (e.g., intersections, campus roads, and mixed-use environments).

\noindent\textbf{Evaluation Metrics.} For NGSIM, HighD, and MoCAD, we report Root Mean Square Error (RMSE) to quantify trajectory accuracy over time. In ApolloScape, following official standard protocol, we adopt Average Displacement Error (ADE) and Final Displacement Error (FDE) as evaluation metrics. To account for heterogeneous agent types, we compute a weighted sum of ADE and FDE across vehicles, pedestrians, and bicycles. Formally,
\begin{equation}
\begin{aligned}
\text{WSADE} &= D_v \cdot \text{ADE}_v + D_p \cdot \text{ADE}_p + D_b \cdot \text{ADE}_b, \\
\text{WSFDE} &= D_v \cdot \text{FDE}_v + D_p \cdot \text{FDE}_p + D_b \cdot \text{FDE}_b,
\end{aligned}
\end{equation}
where \( D_v = 0.20 \), \( D_p = 0.58 \), and \( D_b = 0.22 \) represent the relative weights assigned to vehicles, pedestrians, and bicycles, respectively. These metrics allow for a comprehensive and fair comparison across heterogeneous traffic agents and ensure that model performance reflects safety-critical forecasting across all modalities.}

{
\subsection{Implementation and Training Details}
\subsubsection{Training Details}
We conduct the coding and experiments on a device equipped with an NVIDIA GeForce 3090 GPU.  We train our model using the Adam optimizer and adopt the CosineAnnealingWarmRestarts strategy for dynamic learning rate scheduling. The initial learning rate is set to $lr = 0.001$, with a batch size of $B = 128$. During training, each input segment is split with a factor of 2 to increase temporal resolution. To reduce the adverse impact of misclassifying maneuver types on motion forecasting accuracy and robustness, we augment the MSE criterion with the NLL criterion in the loss function. Specifically, the proposed loss function $\mathcal{L}$ combines two components: Mean Squared Error (MSE) loss $\mathcal{L}_{\textit{MSE}}$ and Negative Log-Likelihood (NLL) loss $\mathcal{L}_{\textit{NLL}}$.
These loss functions collectively guide the training process, ensuring that the model accurately forecasts motion and aligns with real-world driving maneuvers.

\subsubsection{Data-driven RSS Parameter Examination}
To enhance safety awareness, we embed environment-specific parameters from the RSS framework directly into the training pipeline, distinguishing between highway and urban scenarios. For highways, we use a lateral fluctuation margin of $1.0\, \text{m}$ and a reaction time of $\rho = 0.8\, \text{s}$, while urban settings adopt tighter constraints, including a reduced safety buffer of $2.5\, \text{m}$. The safe longitudinal proximity threshold is set to \( d_{\textit{close}}^{\textit{lon}} = 2.0 \, \text{m} \). mportantly, acceleration and deceleration capabilities—both longitudinal and lateral—vary across different traffic contexts due to road surface conditions (e.g., dry, wet), weather (e.g., clear, rainy), vehicle properties (e.g., mass, drivetrain), driving styles (e.g., aggressive, conservative), time of day, and scene type (e.g., urban, highway). Rather than using static thresholds~\cite{naumann2021responsibility}, we employ a data-driven parameter estimation strategy. Specifically, RSS-relevant parameters such as maximum acceleration, braking limits, and lateral maneuverability are dynamically inferred from historical trajectories of traffic agents.
This approach enables the model to adaptively capture scene-specific safety constraints grounded in real-world driving behavior, improving realism and reducing over-conservatism. By learning from empirical patterns, the model can better reflect varying traffic densities, agent types, and risk tolerances. Then, these safety-aware parameters are incorporated via the DSG, ensuring physics-consistent safety modeling across diverse and complex driving scenarios.

\subsubsection{Implementation Details}
We provide further details on the implementation and important parameter settings as follows: 

\noindent\textbf{Intention-aware Module:} This module constructs the Dynamic Interaction Graph \( \mathcal{G}_{\textit{DIG}}^{t-t_{h}:t} \), following the edge construction strategy of GRIP \citep{li2019grip} with an interaction radius \( d_{\textit{close}} = 25 \). It is then processed by an uncertainty-aware GAT, where the number of attention heads \( K_{\textit{GAT}} \) is set to 2, and the mean variance of the learnable Gaussian noise, introduced via the GUF, is initialized to 0.

\noindent\textbf{Safety-aware Module:} This module consists of a spatial encoder and a temporal encoder. The spatial encoder builds the DSG \( \mathcal{G}_{\textit{DSG}}^{t-t_{h}:t} \) based on RSS metrics and processes it using the same uncertainty-aware GAT. The temporal encoder employs an LSTM with a hidden size of \( \mathcal{C} = 64 \), capturing temporal dependencies in the safety-aware features.

\noindent\textbf{Feature Fusion Module:} This module fuses three feature types: \( \mathcal{F}_I \) (intention-aware), \( \mathcal{F}_S \) (safety-aware), and \( \mathcal{F}_T \) (temporal). Feature integration uses a Conv2D operation with kernel size 3 and padding 1, as described in Eq.~\ref{eq:conv}, and a multi-head dot-product attention mechanism with 4 heads.

\noindent\textbf{Multi-modal Decoder:} This component introduces maneuver-aware multimodality to support diverse motion outcomes. It uses an LSTM with a hidden size of 64 to produce a multimodal distribution over future trajectories.

\noindent\textbf{SafeCast (small):} We propose a lightweight variant of our model, SafeCast (small), which omits the temporal encoder from the safety-aware module and merges the intention-aware and safety-aware components into a unified uncertainty-aware GAT. The input node set is redefined as \( \mathcal{V}^{t_{k}} = \{ v_{i}^{t_k} \mid t_{k} \in [t-t_{h}, t], \, i \in [0, n] \} \), where each node \( v_{i}^{t_k} \) includes \( \{ \bm{X}_{i}^{t_{k}}, d_{\textit{max}, i}^{\textit{lon}, t_k}, d_{\textit{max}, i}^{\textit{lat}, t_k} \} \). The resulting representation \( F_{IS} \in \mathbb{R}^{\mathcal{B} \times t_h \times n \times \mathcal{C}} \) is projected to \( F_T \in \mathbb{R}^{\mathcal{B} \times t_h \times \mathcal{C}} \) by selecting the 0-th index along the agent dimension, corresponding to the ego vehicle. This compact variant retains core capabilities while significantly improving computational efficiency.

}

\subsection{Evaluation Results}
\subsubsection{Compare to SOTA Baselines}

\begin{table*}[tb]
\centering
\caption{Evaluation results in the ApolloScape dataset. $\textit{ADE}_{v/p/b}$ and $\textit{FDE}_{v/p/b}$ are the ADE and FDE metrics for the vehicles, pedestrians, and bicycles, respectively. \textbf{Bold} and \underline{underlined} values represent the best and second-best performance in each category.}
\resizebox{0.95\linewidth}{!}{
\begin{tabular}{c|c|c|c|ccc|c|ccc}
\bottomrule
Method & Publication& Backbone & WSADE & ADEv & ADEp & ADEb & WSFDE & FDEv & FDEp & FDEb \\
\hline
TrafficPredict \cite{ma2019trafficpredict} & {AAAI'19} & LSTM & 8.5881 & 7.9467 & 7.1811 & 12.8805 & 24.2262 & 12.7757 & 11.1210 & 22.7912 \\
TPNet \cite{fang2020tpnet} & {CVPR'20} & CNN & 1.2800 & 2.2100 & 0.7400 & 1.8500 & 2.3400 & 3.8600 & 1.4100 & 3.4000 \\
S2TNet \cite{chen2021s2tnet}  & {ACML'21} & Transformer & 1.1679 & {1.9874} & 0.6834 & 1.7000 & 2.1798 & 3.5783 & 1.3048 & 3.2151 \\
MSTG \cite{mstg} & {IET-ITS'23} & LSTM & 1.1546 & \underline{1.9850} & 0.6710 & \underline{1.6745} & \underline{2.1281} & 3.5842 & \underline{1.2652} & \underline{3.0792} \\
TP-EGT \cite{tp-egt} & {IEEE-TITS'24} & Transformer & 1.1900 & 2.0500 & 0.7000 & 1.7200 & 2.1400 & 3.5300 & 1.2800 & 3.1600 \\
\hline
\textbf{SafeCast} & {--} & GAT + GUF & \textbf{1.1253} & \textbf{1.9372} & \textbf{0.6561} & \textbf{1.6247} & \textbf{2.1024} & \textbf{3.5061} & \textbf{1.2524} & \textbf{3.0657} \\
\toprule
\end{tabular}
}
\label{tab:apollo}
\end{table*}

\begin{table}[t]
  \centering
  \caption{{Evaluation results for our proposed model and other SOTA baselines on the NGSIM, HighD, and MoCAD datasets across various prediction horizons. The evaluation metric is RMSE (m), where lower values indicate better performance. Some results are unspecified (``-''). \textbf{Bold} values denote the best performance, and \underline{underlined} values indicate the second-best performance in each category. ``AVG'' represents the average RMSE across all horizons.}}
   \setlength{\tabcolsep}{2.5mm}
   \resizebox{\linewidth}{!}{
    \begin{tabular}{c|ccccccc}
    \toprule
    \multicolumn{1}{c}{\multirow{2}[4]{*}{Dataset}} & \multirow{2}[4]{*}{Model} & \multicolumn{6}{c}{Prediction Horizon (s)} \\
\cmidrule{3-8}    \multicolumn{1}{c}{} &       & 1     & 2     & 3     & 4     & 5  &AVG\\
    \midrule
    \multirow{15}[3]{*}{NGSIM} 
         & S-LSTM \citep{alahi2016social} & 0.65  & 1.31  & 2.16  & 3.25  & 4.55  & 2.38  \\
                   & M-LSTM \citep{Deo_2018} & 0.58  & 1.26  & 2.12  & 3.24  & 4.66  & 2.37  \\
          & CS-LSTM \citep{deo2018convolutional} & 0.61  & 1.27  & 2.09  & 3.10  & 4.37  & 2.29  \\
          & MATF-GAN \citep{zhao2019multi} & 0.66  & 1.34  & 2.08  & 2.97  & 4.13  & 2.24  \\
          & DRBP \citep{gao2023dual} & 1.18  & 2.83  & 4.22  & 5.82  & - & 3.51  \\
          & MFP \citep{tang2019multiple} & 0.54  & 1.16  & 1.89  & 2.75  & 3.78  & 2.02  \\
          & NLS-LSTM \citep{messaoud2019non} & 0.56  & 1.22  & 2.02  & 3.03  & 4.30  & 2.23  \\
          & MHA-LSTM \citep{messaoud2020attention} & 0.41  & 1.01  & 1.74  & 2.67  & 3.83  & 1.93  \\  
          & WSiP \citep{Wang_Wang_Yan_Wang_2023} & 0.56  & 1.23  & 2.05  & 3.08  & 4.34  & 2.25  \\  
          & CF-LSTM \citep{xie2021congestion} & 0.55  & 1.10  & 1.78  & 2.73  & 3.82  & 2.00  \\  
          & STDAN \citep{chen2022intention} & 0.42  & 1.01  & 1.69  & 2.56  & 3.67  & 1.87  \\  
          & DACR-AMTP \citep{cong2023dacr} & 0.57  & 1.07  & 1.68  & 2.53  & \underline{3.40}  & 1.85  \\ 
          & HTPF \citep{hybrid} & 0.49  & 1.09  & 1.78  & 2.62  & 3.65  & 1.92 \\ 
            & BAT \cite{liao2024bat} &\textbf{0.23} & \textbf{0.81} & \underline{1.54} & \underline{2.52} & {3.62} & \underline{1.74} \\
          & FHIF \citep{zuo2023trajectory} & \underline{0.40}  & 0.98  & 1.66  & \underline{2.52}  & 3.63  & 1.84  \\  
          & \textbf{SafeCast} & \underline{0.40}  & \underline{0.90}  & \textbf{1.42}  & \textbf{2.05}  & \textbf{2.89}  & \textbf{1.54}  \\ 
    \midrule
    \multirow{11}[2]{*}{HighD}
          & S-GAN \citep{gupta2018social}& 0.30  & 0.78  & 1.46  & 2.34  & 3.41  & 1.66  \\ 
          & CS-LSTM \citep{deo2018convolutional}& 0.22  & 0.61  & 1.24  & 2.10  & 3.27  & 1.49  \\ 
          & DRBP\citep{gao2023dual} & 0.41  & 0.79  & 1.11  & 1.40  & - & 0.93  \\ 
          & NLS-LSTM \citep{messaoud2019non}& 0.20  & 0.57  & 1.14  & 1.90  & 2.91  & 1.34  \\ 
          & MHA-LSTM \citep{messaoud2020attention}& 0.19  & 0.55  & 1.10  & 1.84  & 2.78  & 1.29  \\ 
          & HTPF \citep{hybrid} & 0.16  & 0.47  & 0.94  & 1.58  & 2.36  & 1.10 \\ 
          & STDAN \citep{chen2022intention}& 0.19  & 0.27  & 0.48  & 0.91  & 1.66  & 0.70  \\ 
          & iNATran \citep{chen2022vehicle}& \textbf{0.04}  & \textbf{0.05}  & \underline{0.21}  & 0.54  & 1.10  & \underline{0.39}  \\ 
          & DACR-AMTP \citep{cong2023dacr}& 0.10  & 0.17  & 0.31  & 0.54  & \underline{1.01}  & 0.43  \\
            &GaVa \cite{liao2024human} & 0.17  & 0.24  & 0.42  & 0.86  & \underline{1.31} &0.60  \\ 
          & \textbf{SafeCast (small)} & 0.24  & 0.39  & 0.57  & 0.70  & 1.08  & 0.58  \\ 
          & \textbf{SafeCast} & \underline{0.06}  & \underline{0.10}  & \textbf{0.14}  & \textbf{0.35}  & \textbf{0.79}  & \textbf{0.29}  \\  
    \midrule
    \multirow{8}[2]{*}{MoCAD} 
          & CS-LSTM \citep{deo2018convolutional}& 1.45  & 1.98  & 2.94  & 3.56  & 4.49 & 2.88  \\ 
          & MHA-LSTM \citep{messaoud2020attention}& 1.25  & 1.48  & 2.57  & 3.22  & 4.20  & 2.54  \\ 
          & NLS-LSTM \citep{messaoud2019non}& 0.96  & 1.27  & 2.08  & 2.86  & 3.93  & 2.22  \\ 
          & CF-LSTM \citep{xie2021congestion}& 0.72  & 0.91  & 1.73  & 2.59  & 3.44  & 1.88  \\ 
          & WSiP \citep{Wang_Wang_Yan_Wang_2023}& 0.70  & \underline{0.87}  & 1.70  & 2.56  & 3.47  & 1.86  \\ 
          & STDAN \citep{chen2022intention}& \underline{0.62} & \textbf{0.85} & \underline{1.62}  & \underline{2.51} & \underline{3.32}  & \underline{1.78}  \\
            
          & \textbf{SafeCast (small)} & 0.70  & 1.20  & 1.87  & 2.60  & 3.42  & 1.96  \\ 
          & \textbf{SafeCast} & \textbf{0.50}  & \textbf{0.85}  & \textbf{1.57}  & \textbf{2.19}  & \textbf{2.99 }& \textbf{1.62}  \\  
    \bottomrule
    \end{tabular}%
    }
  \label{tab:benchmark}%
\end{table}%

{ Table~\ref{tab:benchmark} and Table~\ref{tab:apollo}  summarizes the performance of SafeCast against state-of-the-art (SOTA) baselines across four benchmark datasets: NGSIM, HighD, ApolloScape, and MoCAD. The results demonstrate that SafeCast consistently achieves superior accuracy, outperforming existing methods in both short-term and long-term motion forecasting tasks.

On the NGSIM dataset, SafeCast delivers substantial gains over earlier baselines (2017–2020), with improvements of at least 10.89\% for short-term and 23.14\% for long-term predictions. Compared to more recent methods (2021–2024), it maintains a strong lead, reducing 5-second RMSE by a minimum of 0.51 meters and achieving at least 15.00\% improvement in long-horizon accuracy. These enhancements are particularly significant for reducing accident risk in complex, high-speed driving scenarios. Correspondingly, in the HighD dataset, although short-term predictions are competitive across models, SafeCast shows marked improvements in long-term accuracy, with a minimum RMSE reduction of 21.78\%. Notably, even the lightweight SafeCast (small) variant outperforms most full-scale models, demonstrating the efficiency and scalability of our SafeCast model.

Moreover, on the ApolloScape dataset, a challenging urban dataset with diverse agent types (vehicles, pedestrians, and cyclists), SafeCast achieves the best overall performance. As shown in Table~\ref{tab:apollo}, it records a WSADE of 1.1253 and WSFDE of 2.1024, outperforming the previous best baseline (MSTG) by 2.5\% and 1.2\%, respectively. It also establishes new benchmarks across all agent-specific ADE and FDE metrics, underscoring its effectiveness in modeling fine-grained, heterogeneous interactions. The integration of uncertainty-aware GAT and Global Uncertainty Fusion proves especially beneficial for dynamic agents like pedestrians and cyclists. Finally, in the urban, multi-agent environments of the MoCAD dataset, SafeCast maintains its leading performance. It achieves an average RMSE improvement of at least 8.90\% over competing methods and reduces long-term RMSE by 0.32–0.33 meters. These results further validate SafeCast’s capacity to handle dense, interactive traffic scenes with high temporal variability.

Overall, SafeCast demonstrates outstanding generalization and robustness across diverse driving scenarios, consistently advancing the state of the art in both accuracy and reliability for motion forecasting in real-world AD systems.}

\begin{figure}[t]
\centering
\includegraphics[width=0.40\textwidth]{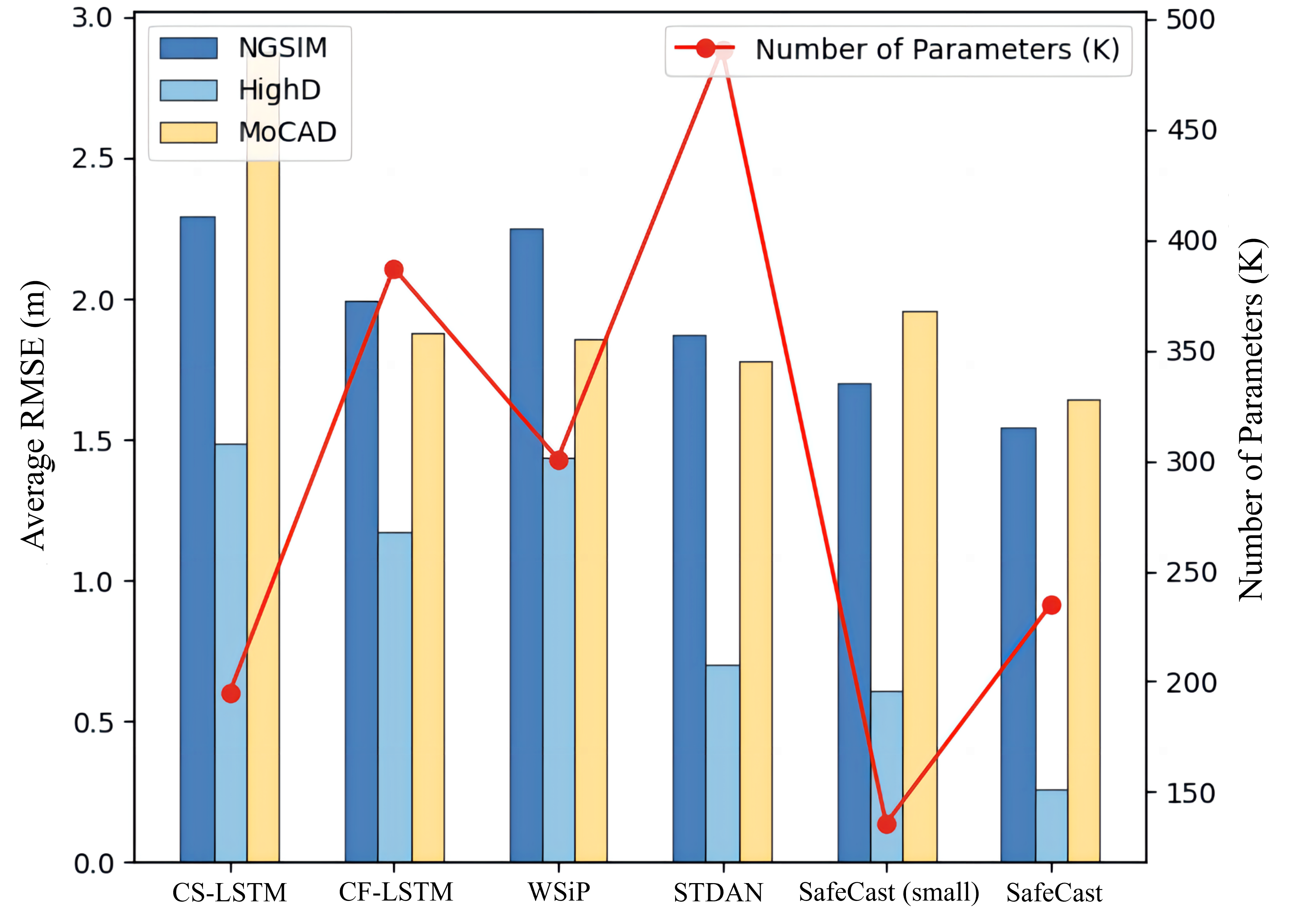}
\caption{Comparative analysis of the performance of the models and their parametric counts.}
\label{fig:param_1}
\end{figure}

\begin{table}[htbp]
  \centering
  \caption{Comparative evaluation of our proposed models with selected baselines. Highlighting the prediction accuracy metric (Average RMSE) and model complexity measured by the number of parameters (\#Param.). \textbf{Bold} and \underline{underlined} values represent the best and second-best performance.}
  \resizebox{0.85\linewidth}{!}{
    \begin{tabular}{ccccccc}
    \bottomrule
    \multirow{2}[4]{*}{Model} & \multicolumn{3}{c}{Average RMSE (m)} & \multirow{2}[2]{*}{\#Param. (K)}\\
\cmidrule{2-4}          & NGSIM     & HighD   &MoCAD   \\
    \midrule
    CS-LSTM \citep{deo2018convolutional} & 2.29 & 1.49 &2.88 & \underline{194.92}\\
    CF-LSTM \citep{xie2021congestion}& 1.99 & 1.17 &1.88 & 387.10 \\
    WSiP \citep{Wang_Wang_Yan_Wang_2023}& 2.25 & 1.44 &1.86 & 300.76\\
    STDAN \citep{chen2022intention}& 1.87 & 0.70 &\underline{1.78} & 486.22 \\
    SafeCast (small) & \underline{1.64} & \underline{0.58} &1.96 & \textbf{135.83} \\
    SafeCast & \textbf{1.54} & \textbf{0.29}  &\textbf{1.62} &  234.55\\
   \toprule 
    \label{tab:param}
\end{tabular}}%
\end{table}%

\begin{table}[htbp]
  \centering 
  \caption{{Performance comparison of proposed models with the SOTA baselines on NGSIM dataset. The batch size for evaluation is set to 128. A total of 10 batches are predicted, and the average inference time (s) and average RMSE (m) of the predictions are calculated for each baseline.}}
  \resizebox{0.85\linewidth}{!}{
    \begin{tabular}{ccccccc}
        \bottomrule
    {Model} & {Average RMSE (m)} &{Inference time (s)}\\
    \midrule
    M-LSTM \citep{Deo_2018}& 2.37 & \textbf{0.14}\\
    S-LSTM \citep{alahi2016social}& 2.38  & 0.30 \\
    CS-LSTM \citep{deo2018convolutional} & 2.29 & 0.37\\
    MHA-LSTM \citep{messaoud2020attention}& 1.93  & 0.23 \\
    TS-GAN \citep{wang2020multi}& 2.06  & 0.39 \\
    DACR-AMTP \citep{cong2023dacr}& 1.85 & 0.26 \\
    STDAN \citep{chen2022intention}& 1.87 & 0.21 \\
    SafeCast (small) & \underline{1.70} & \underline{0.17}\\
    SafeCast & \textbf{1.54}  &  0.20\\
  \toprule
    \label{tab:infer}
\end{tabular}}
\end{table}%

\subsubsection{Inference Speed and Model Complexity}
\begin{figure}
\centering
\includegraphics[width=0.4\textwidth]{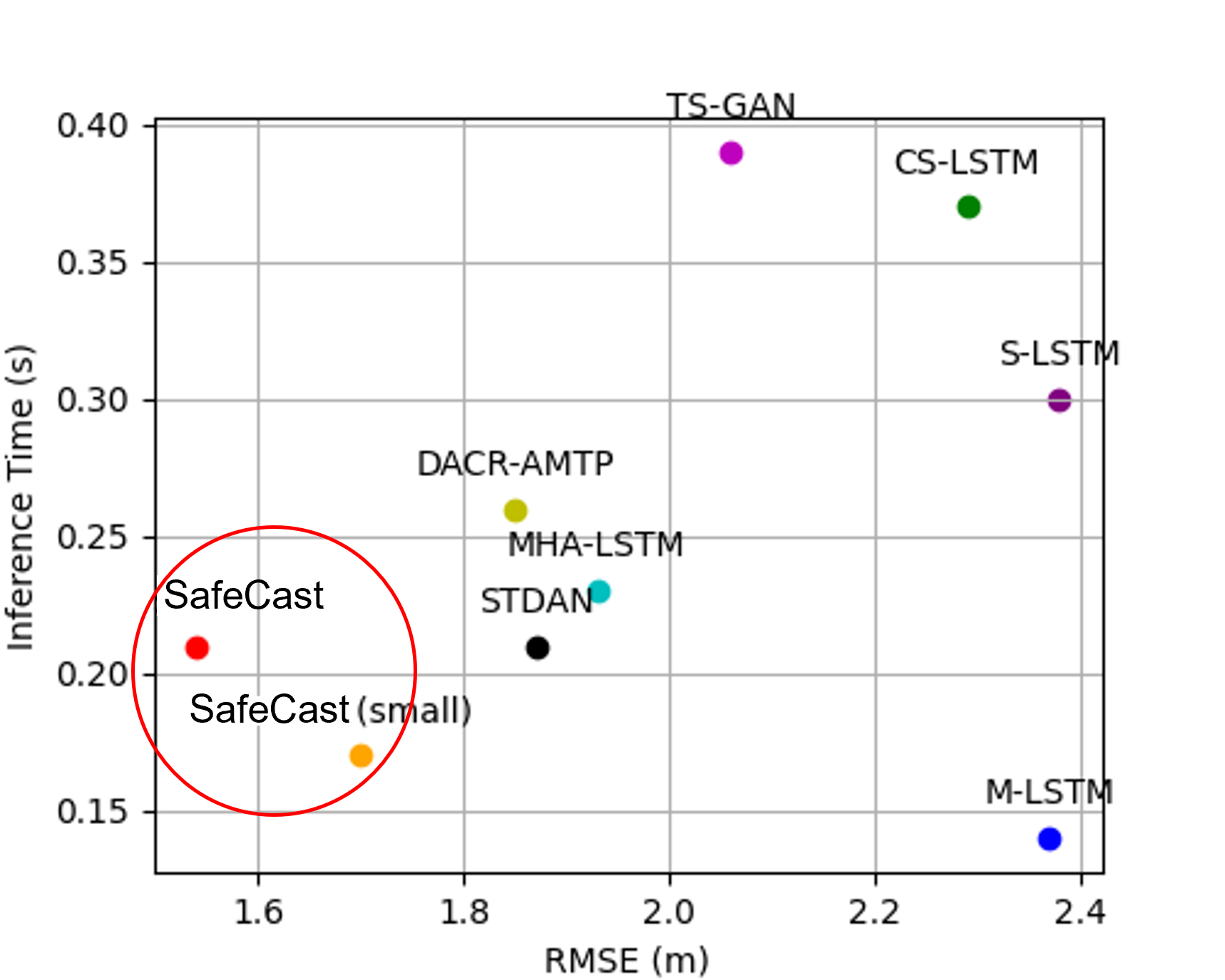}
\caption{Inference speed versus prediction accuracy comparison between SafeCast and open-source baselines. Models closer to the origin achieve superior performance. The top two performers are marked in a red circle.}
\label{fig:infer}
\end{figure}

To further demonstrate the scalability and efficiency of our SafeCast and SafeCast (small), we conduct a detailed benchmark analysis of their inference speed and complexity against other SOTA models. The results are presented in Table \ref{tab:infer} and Fig. \ref{fig:infer}. Notably, the efficiency metrics are not commonly provided in this field, and our access to the source code of various models is limited. Therefore, our comparison primarily focuses on available open-source models.

\begin{table}[htbp]
  \centering
  \caption{Different methods and components of ablation study.}\label{tab:ablation_description}
  \resizebox{0.95\linewidth}{!}{
    \begin{tabular}{cccccccc}
    \toprule
    \multirow{2}[4]{*}{Components} & \multicolumn{7}{c}{Ablation methods} \\
\cmidrule{2-8}          & A     & B     & C     & D     & E &F &G\\
    \midrule
    Intention-aware Encoder & \ding{56} & \ding{52} & \ding{52} & \ding{52} & \ding{52} & \ding{52} & \ding{52}\\
    Safety-aware Spatial Encoder & \ding{52} & \ding{56} & \ding{52}  & \ding{52} & \ding{52} & \ding{52} & \ding{52}\\
    Safety-aware Temporal Encoder  & \ding{52} & \ding{52} & \ding{56} & \ding{52}  & \ding{52} & \ding{52} & \ding{52}\\
    Multi-modal Decoder & \ding{52} & \ding{52} & \ding{52} & \ding{56}  & \ding{52} & \ding{52} & \ding{52}\\
    Multi-head Attention Mechanism & \ding{52} & \ding{52} & \ding{52} & \ding{52}  & \ding{56} & \ding{52} & \ding{52}\\
    Graph Uncertainty Feature Method & \ding{52} & \ding{52} & \ding{52} & \ding{52}  & \ding{52} & \ding{56} & \ding{52}\\
    \bottomrule
\end{tabular}}%
\end{table}%

{
When compared to the fastest inference model, M-LSTM, SafeCast and SafeCast (small) exhibit slightly slower speeds. However, they demonstrate superior performance on the NGSIM dataset, with an average RMSE 35.02\% higher than M-LSTM. Impressively, SafeCast and SafeCast (small) improve inference speed by at least 4.76\% and 19.05\%, respectively, compared to other SOTA models, while ensuring better prediction performance.  The inference speed is highly dependent on the specific algorithm implementation and the computing hardware used, and it can vary significantly even for identical model architectures. Moreover, the parameter analysis depicted in Table \ref{tab:param} and Fig. \ref{fig:param_1} illustrates that SafeCast achieves high precision in motion forecasting with a modest parameter quantity (just 234.55K). This figure represents a reduction of 39.41\% and 22.01\%, compared to the leading baselines CF-LSTM and WSiP, respectively. Notably, SafeCast (small) has the fewest parameters while outperforming most of the state-of-the-art baseline models regarding prediction accuracy. These findings underscore the efficient and streamlined design of SafeCast (small), highlighting its prowess in motion forecasting.}

\subsection{Ablation Study}

\begin{table*}[!htbp]
\centering
\caption{Ablation results for different models. \textbf{Bold} values represent the best performance. (Evaluation metric: RMSE (m))}\label{tab:ablation_comparison}
\setlength{\tabcolsep}{3mm}
\resizebox{0.9\linewidth}{!}{
\begin{tabular}{cccccccc}

\toprule
\makecell[c]{Prediction\\Horizon (s)}
&\makecell[c]{Method A\\($\Delta${Method} G)}
&\makecell[c]{Method B\\($\Delta${Method} G)}
&\makecell[c]{Method C\\($\Delta${Method} G)} 
&\makecell[c]{Method D\\($\Delta${Method} G)}
&\makecell[c]{Method E\\($\Delta${Method} G)}
&\makecell[c]{Method F\\($\Delta${Method} G)}
&\makecell[c]{Method G\\(Full model)}\\

\midrule
1 & 0.42$_{\downarrow{5.00\%}}$ & 0.42$_{\downarrow{5.00\%}}$ &0.45$_{\space\downarrow{12.50\%}}$ &0.43$_{\downarrow{7.50\%}}$ &\textbf{0.40}$_{\space\downarrow{0.00\%}}$ &0.42$_{\downarrow{5.00\%}}$ &\textbf{0.40}\\
2 & 0.96$_{\downarrow{6.67\%}}$ & 0.94$_{\downarrow{4.44\%}}$ &0.95$_{\downarrow{5.56\%}}$ &1.02$_{\downarrow{13.33\%}}$ &0.92$_{\downarrow{2.22\%}}$ &0.94$_{\downarrow{4.44\%}}$ &\textbf{0.90}\\
3 & 1.56$_{\downarrow{9.86\%}}$ & 1.49$_{\downarrow{4.93\%}}$ &1.49$_{\downarrow{4.93\%}}$ &1.71$_{\downarrow{20.42\%}}$ &1.45$_{\downarrow{2.11\%}}$ &1.51$_{\downarrow{6.34\%}}$ &\textbf{1.42}\\
4 & 2.27$_{\downarrow{10.73\%}}$ & 2.15$_{\downarrow{4.88\%}}$ &2.12$_{\downarrow{3.41\%}}$ &2.59$_{\downarrow{26.34\%}}$ &2.08$_{\downarrow{1.46\%}}$ &2.13$_{\downarrow{3.90\%}}$ &\textbf{2.05}\\
5 & 3.22$_{\downarrow{11.42\%}}$ & 3.02$_{\downarrow{4.50\%}}$ &2.96$_{\downarrow{2.42\%}}$ &3.70$_{\downarrow{28.03\%}}$ &2.93$_{\downarrow{1.38\%}}$ &3.01$_{\downarrow{4.15\%}}$ &\textbf{2.89}\\

\bottomrule
\end{tabular}
}
\end{table*}

\begin{figure}
\centering
\includegraphics[width=0.45\textwidth]{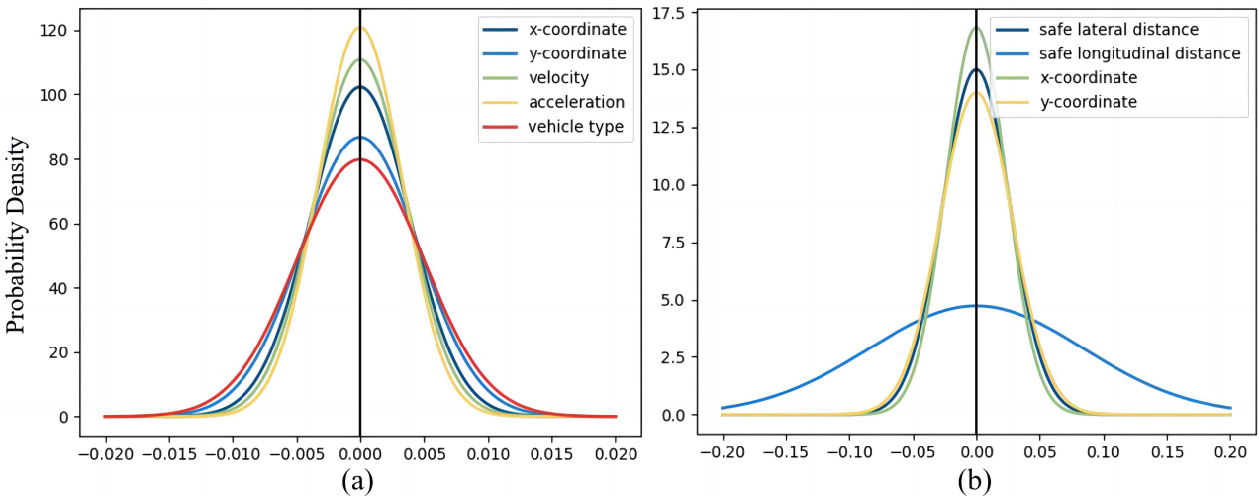}
\caption{The visualization shows the variable-learnable noise distribution produced by the GUF layer, illustrating how the noise adapts to different contexts: (a) intention-aware module, (b) safety-aware module.}
\label{fig:distribution}
\end{figure}

{
To evaluate the individual contributions of each component in our model, we conduct a series of ablation studies. Table \ref{tab:ablation_description} outlines the specific modifications for each method: Method A excludes the Intention-aware module; Method B omits the spatial encoder from the Safety-aware module, while Method C removes the temporal encoder; Method D eliminates the multimodal probabilistic maneuver prediction in the decoder; Method E replaces the multi-head attention mechanism with a convolutional neural network; Method F excludes the GUF method; and Method G includes all components as the complete model.

The ablation results on the NGSIM dataset, shown in Table \ref{tab:ablation_comparison}, highlight the importance of each component in improving model performance. All models were fully trained, and the results reflect their best performance on the test set. Excluding the intention-aware module (Method A) leads to a performance drop of at least 10.73\% in long-term predictions, while removing safety-aware features (Method B) results in a 4.74\% decrease compared to the full model (Method G). Method D exhibits the most substantial performance decline, with a reduction of at least 7.50\%, underscoring the significance of multimodal probabilistic maneuver prediction in the decoder. Furthermore, incorporating the temporal encoder within the Safety-aware module proves essential for capturing the temporal dependencies of the target vehicle. The comparison between Method C (without the temporal encoder) and Method G highlights its importance, particularly for short-term prediction accuracy.  The benefits of the GUF method in handling uncertainties are evident in Method F, where excluding GUF results in noticeable performance deterioration. Finally, replacing the multi-head attention mechanism with a convolutional neural network in Method E causes a performance decrease of at least 4.2\%. These findings emphasize the critical role of each component in motion forecasting, as demonstrated by the superior results of the Method G.}

{\subsection{Qualitative Results}

\subsubsection{Qualitative Results of GUF Noise Distribution}
We integrate the GUF into both the intention-aware module and the spatial encoder of the safety-aware module. This design addresses a core challenge in modeling traffic dynamics: the lack of universally accepted standards for noise distribution in historical state elements—such as velocity, acceleration, and 2D positional coordinates—within the same traffic scenario. Instead of relying on fixed noise thresholds, GUF introduces learnable Gaussian noise parameters that dynamically adapt to the characteristics of the input data. This flexibility allows the prediction model to better capture uncertainty and improve generalization.

To assess the effectiveness of GUF, we visualize the learned noise distributions for both modules in Fig.~\ref{fig:distribution}. These distributions exhibit normal-like behavior, with the range and variance varying across different input features. In Fig.~\ref{fig:distribution}(a), representing the intention-aware module, the learned noise for acceleration differs from that of the \( y \)-coordinate, illustrating GUF’s capacity to differentiate uncertainty based on the relational significance of each feature. The relatively compact distributions in this module are likely due to the strong numerical and semantic correlations among position, velocity, and acceleration.

In contrast, Fig.~\ref{fig:distribution}(b) shows the learned noise distributions in the spatial encoder, which incorporates features such as coordinates and safety distances. These features have weaker interdependencies compared to the intention-aware module, resulting in broader and more diverse noise ranges. This suggests that the spatial encoder captures more latent, abstract safety representations beyond conventional physical dynamics. Overall, the GUF enables the model to flexibly adapt its noise modeling strategy to reflect the underlying complexity of intention and safety features, supporting more robust and context-aware prediction.

\begin{figure*}
\centering
\includegraphics[width=0.8\textwidth]{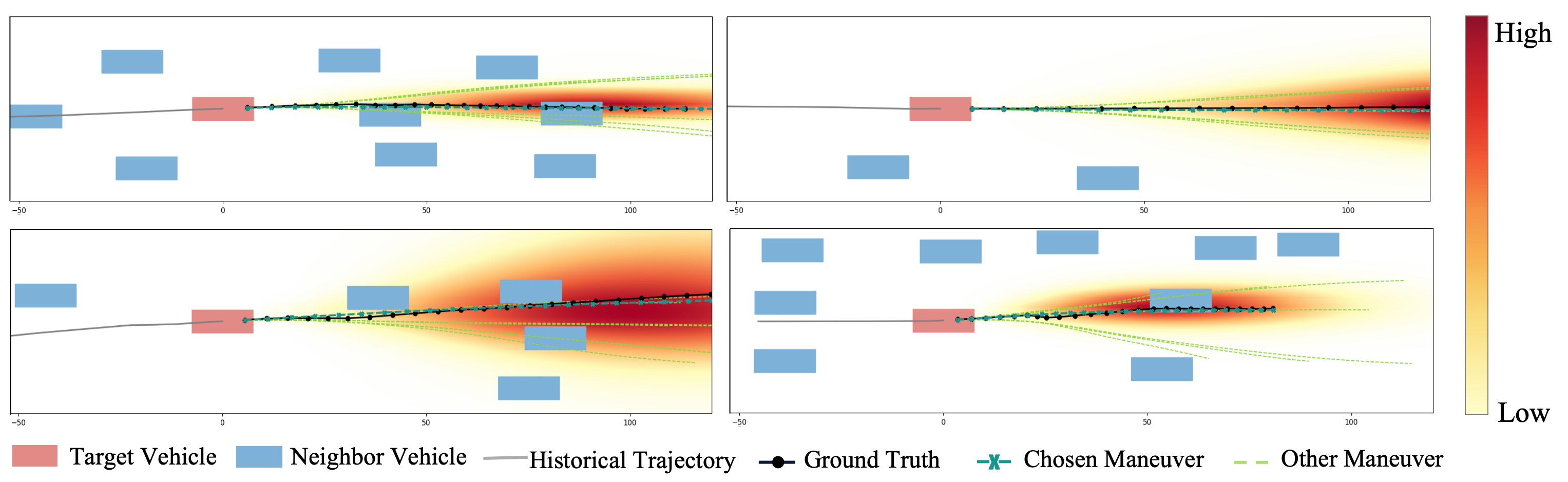}
\caption{Multi-modal probabilistic motion forecasting of the target vehicle. Heat maps show the Gaussian Mixture Model (GMM) of the predictions, with brighter regions indicating higher probabilities.}
\label{fig:multimodal}
\end{figure*}

\begin{figure*}
\centering
\includegraphics[width=0.8\textwidth]{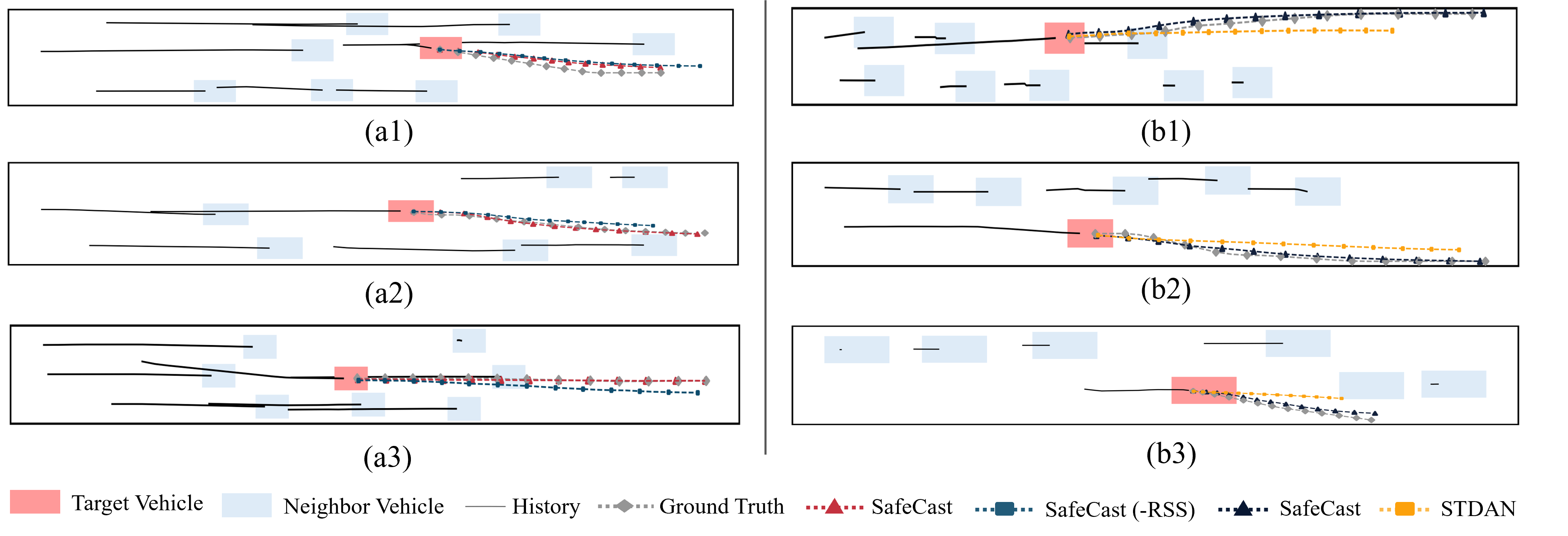}
\caption{Qualitative results of our proposed models on the NGSIM dataset. (a1)-(a3): Visualization results of motion forecasting between SafeCast and SafeCast (-RSS) with and without RSS metrics; (b1)-(b3): Comparison of visualization results between SafeCast and STDAN. The target vehicle is depicted in red, while its surrounding vehicles are shown in light blue.}
\label{fig:visualization}
\end{figure*}

\subsubsection{Qualitative Results for Multimodal Forecasting}
Fig. \ref{fig:multimodal} illustrates the multimodal probabilistic forecasting performance of the SafeCast model on the NGSIM dataset. The heatmaps depict the GMM's predictions at each time step, with brighter colors indicating higher probabilities for the forecasted motion. These results confirm that the most probable forecasts closely match the ground truth, demonstrating the impressive performance of SafeCast.

\subsubsection{Qualitative Results for Real-world Datasets}

We present qualitative visualizations of SafeCast's motion forecasting in real-world safety-critical scenarios, with a focus on the impact of incorporating the Responsibility-Sensitive Safety (RSS) metric. Safety-critical scenarios refer to situations where collision avoidance and maintaining safe distances are essential—such as high-density traffic, sudden deceleration of leading vehicles, or abrupt lane changes. To evaluate the effect of integrating RSS, we compare the performance of SafeCast with and without safety constraints, as illustrated in Fig.~\ref{fig:visualization} (a1), (a2), and (a3).

In Fig.~\ref{fig:visualization} (a1), the target vehicle is closely surrounded by other vehicles in the front and adjacent lanes. SafeCast, when incorporating the RSS metric, calculates both longitudinal and lateral safety distances, prompting the vehicle to decelerate appropriately and maintain a safe buffer, resulting in a predicted trajectory closely aligned with the ground truth. In contrast, SafeCast without the RSS module (SafeCast~(-RSS)) disregards these constraints and fails to model safe spacing, leading to significant deviation from the true path.

Fig.~\ref{fig:visualization} (a2) presents a scenario with minimal surrounding traffic, allowing the target vehicle to move more freely. In this case, SafeCast correctly identifies the lack of immediate obstacles and generates an extended motion forecast consistent with the ground truth. Meanwhile, SafeCast~(-RSS) underestimates the available motion space, highlighting its inability to adaptively recognize safe conditions.

Fig.~\ref{fig:visualization} (a3) further demonstrates SafeCast’s handling of lateral safety. When a nearby vehicle is in an adjacent lane, SafeCast correctly infers the importance of lateral separation, predicting a cautious path that preserves lane integrity and avoids potential conflict. Conversely, SafeCast~(-RSS) forecasts an unsafe lane-change maneuver, diverging significantly from the ground truth and exposing a failure in lateral risk awareness.

To further highlight SafeCast’s robustness, we compare it with a state-of-the-art baseline, STDAN, under similar conditions, as shown in Fig.~\ref{fig:visualization} (b1)–(b3). Across various complex scenarios—including lane merging and traffic-dense environments—SafeCast consistently produces forecasts more closely aligned with the ground truth. These visual comparisons underscore the model’s ability to incorporate real-time safety considerations, and its superior reliability in dynamic and uncertain traffic contexts. Overall, these qualitative results demonstrate SafeCast’s adaptability, scene comprehension, and safety-awareness, reinforcing its effectiveness for real-world motion forecasting.}

\section{Conclusion} \label{sec:6}
Accurate motion forecasting of surrounding vehicles is a pivotal challenge in achieving fully autonomous driving. To tackle this challenge, we propose SafeCast, a novel Risk-Responsive motion forecasting model, designed to enhance prediction accuracy while prioritizing safety. This model integrates the Responsibility-Sensitive Safety model to characterize the safety attributes of traffic agents while employing a graph-based framework for real-time modeling of their intentions and safety features. In addition, we propose a variable-learnable noise approach to simulate the inherent uncertainties of traffic environments, which enhances the adaptability and robustness of our model. We also implement a feature fusion strategy to navigate the intricate relationships among various features. Furthermore, we utilize a GMM-based decoder for multimodal prediction to improve the model's forecasting precision.
Evaluated on the NGSIM, HighD, ApolloScape, and MoCAD real-world datasets, SafeCast achieves SOTA performance while maintaining a relatively lightweight architecture and quick inference speeds, demonstrating its robustness, efficiency, and practical applicability in autonomous driving systems.
{\noindent\textbf{Limitations and Future Work.} While SafeCast demonstrates superior performance and practical utility, we recognize several limitations in its current formulation. The design of the RSS-based module prioritizes a trade-off between interpretability, real-world applicability, and computational efficiency. While the data-driven strategy is productive, it may not fully capture the nuanced variability of real-world driving behavior across diverse conditions. In the future, we plan to extend the temporal observation window to better capture long-term behavioral patterns, refine the granularity of RSS parameterization to account for varying driver profiles and vehicle capabilities, and incorporate richer contextual information, such as road surface conditions, weather, and vehicle types. These enhancements are expected to further improve the adaptability, robustness, and fidelity of the safety-aware module in real-world driving scenarios.
}

\section*{Acknowledgement}
This research is supported by Science and Technology Development Fund of Macau SAR (File no. 0021/2022/ITP), State Key Lab of Intelligent Transportation System (2024-B001), Jiangsu Provincial Science and Technology Program (BZ2024055), University of Macau (SRG2023-00037-IOTSC, MYRG-GRG2024-00284-IOTSC), and Shenzhen-Hong Kong-Macau Science and Technology Program Category C (SGDX20230821095159012).

\bibliographystyle{cas_model2_names}
%
\bibliography{cas_refs}

\end{document}